\documentclass[10pt,twocolumn,letterpaper]{article}

\usepackage{wacv}
\usepackage{times}
\usepackage{epsfig}
\usepackage{graphicx}
\usepackage{amsmath}
\usepackage{amssymb}
\usepackage{booktabs}

\def\wacvPaperID{1983} 

\wacvapplicationstrack 

\wacvfinalcopy 

\ifwacvfinal
\usepackage[breaklinks=true,bookmarks=false]{hyperref}
\else
\usepackage[pagebackref=true,breaklinks=true,colorlinks,bookmarks=false]{hyperref}
\fi

\begin{document}

\title{FaRO 2: an Open Source, Configurable Smart City Framework for Real-Time Distributed Vision and Biometric Systems}
\author{Joel Brogan\\ 
{\tt\small broganjr@ornl.gov}\and Nell Barber\\
{\tt\small barbercl@@ornl.gov}\and David Cornett\\
{\tt\small cornettdciii@ornl.gov} \and David Bolme\\
{\tt\small bolmeds@ornl.gov}
}




\maketitle
\thispagestyle{empty}

\begin{abstract}
    Recent global growth in the interest of smart cities has led to trillions of dollars of investment toward research and development. These connected cities have the potential to create a symbiosis of technology and society and revolutionize the cost of living, safety, ecological sustainability, and quality of life of societies on a world-wide scale. Some key components of the smart city construct are connected smart grids, self-driving cars, federated learning systems, smart utilities, large-scale public transit, and proactive surveillance systems. While exciting in prospect, these technologies and their subsequent integration cannot be attempted without addressing the potential societal impacts of such a high degree of automation and data sharing. Additionally, the feasibility of coordinating so many disparate tasks will require a fast, extensible, unifying framework. To that end, we propose FaRO2, a completely reimagined successor to FaRO1, built from the ground up. FaRO2 affords all of the same functionality as its predecessor, serving as a unified biometric API harness that allows for seamless evaluation, deployment, and simple pipeline creation for heterogeneous biometric software. FaRO2 additionally provides a fully declarative capability for defining and coordinating custom machine learning and sensor pipelines, allowing the distribution of processes across otherwise incompatible hardware and networks. FaRO2 ultimately provides a way to quickly configure, hot-swap, and expand large coordinated or federated systems online without interruptions for maintenance. Because much of the data collected in a smart city contains Personally Identifying Information (PII), FaRO2 also provides built-in tools and layers to ensure secure and encrypted streaming, storage, and access of PII data across distributed systems.

\end{abstract}
\begin{figure}
    \centering
    \includegraphics[scale=0.3]{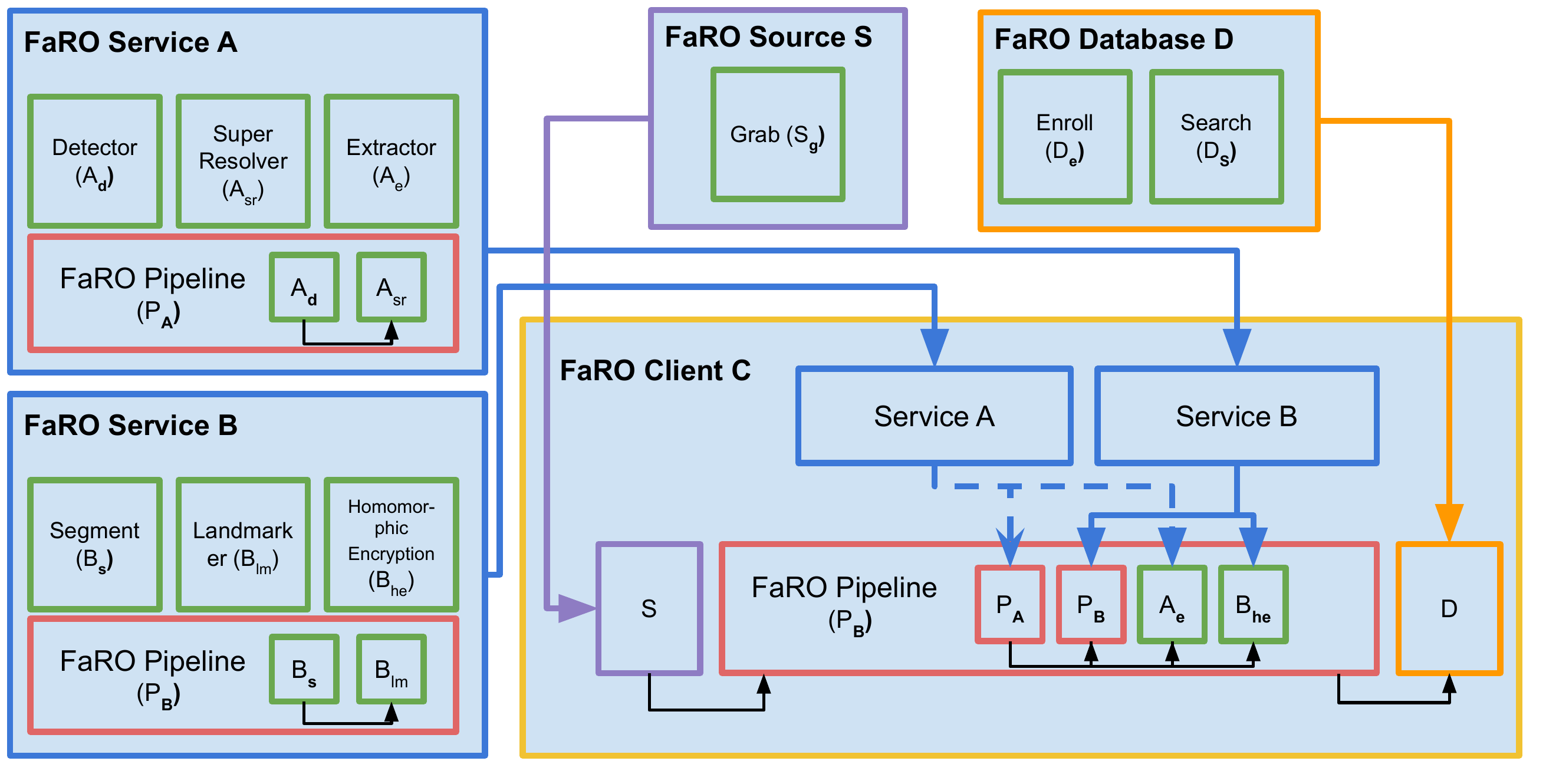}
    \caption{An abstract example of how a declarative FaRO2 computer vision pipelined infrastructure works. }
    \label{fig:faro_arch}
\end{figure}

\section{Introduction}
It is estimated that by 2023, research and development toward smart city applications will reach a market share of over \$700 billion dollars~\cite{khan2020edge}, with hundreds of billions more going towards Internet of Things research. While innovation in this area pushes forward at unprecedented speed, less attention is being directed toward the safe and secure capture and transmission of biometric information the likes of which is vital to an effective smart city implementation. Expanding on the groundwork laid by Face Recognition Oak Ridge (FaRO1) in 2019 \cite{bolme2020face}, FaRO2 is a highly scalable inferencing framework for streaming, processing, and visualization of biometric data across distributed systems. Robust smart city and insight-driven surveillance use cases rely on heterogeneous data sources and software that may be incompatible or prove unwieldy when used in tandem. FaRO2 addresses this problem by providing a unified API harness to accelerate the creation and deployment of custom biometric pipelines, the acquisition of high-quality data, and the capture of real-time insights from streaming video. FaRO2 is also redesigned to include privacy protection and network security components that are critical to its intended applications. Though FaRO2 is primarily configured for tasks related to image-based biometric detection and identification using modalities like face, whole body, and gait, it is flexible enough that data sources capturing different modalities for varied purposes can be integrated with relative ease.

This paper is organized as follows: In Section ~\ref{sec:privacy}, we discuss privacy and potential means by which to protect it with regard to face imagery. In Section ~\ref{sec:related}, we give an overview of related work. In Section ~\ref{sec:architecture}, we provide a technical overview of the FaRO2 architecture. In Section \ref{sec:where}, we provide information about finding FaRO2. In Section \ref{sec:conclusion}, we give a conclusion and discuss future work.

\section{Privacy}
\label{sec:privacy}
Privacy is a key component of smart city research~\cite{kirwan2015defining}; however, at this point in time, no studies, surveys, or software have been published pertaining to this specific topic~\cite{khan2020edge}.

While privacy remains an abstract concept in an era characterized by the confluence of social media, omnipresent digital connectivity, and inescapable data capture, failure to adequately protect biometric information can yield very tangible consequences. Advances in techniques used to reconstruct faces from face feature vectors have necessitated additional safeguards for not just raw face images but also the unique descriptors extracted from them. Reconstructed face images can be submitted to cheap consumer services that can identify other images with matching subjects, many of which contain identifying metadata or are otherwise linked to identifying information online \cite{wenger2022assessing}. This undercuts assurances from many face recognition vendors over the years that, once extracted, these feature vectors do not require encryption because they are already unusable in the wrong hands. 

A popular solution to face feature vector vulnerability is homomorphic encryption (HE), and the application of fully homomorphic encryption (FHE) to biometric matching has been a popular area of research in recent years. HE enables the encryption of face feature vectors such that matching is done in homomorphic space, and plain face feature vectors need never be exposed to networks or to server environments. In spite of its advantages, however, FHE does not support real-time biometric identification and, as it stands, is not a fit for FaRO2. While authors in \cite{8698601} show that simple matching can be performed in real time using FHE, there is little utility in these results when considering the need to encrypt all new faces captured by a particular sensor and search a gallery of arbitrary length for matching individuals. Instead, FaRO2 leverages partially homomorphic encryption (PHE) to serve in its place. The principal distinction between FHE and the PHE utilized by FaRO2 is that the former supports the evaluation of arbitrary functions by way of its support for both addition and multiplication in homomorphic space \cite{acar_aksu_uluagac_conti_2019} while the latter does not. The PHE implementation used by FaRO2 is homomorphic over addition but supports only multiplication between encrypted and unencrypted operands to yield encrypted results. This means that cosine similarity is not supported in homomorphic space and matching must be performed by other means. FaRO2's PHE implementation is described in greater technical detail in Section~\ref{sec:security}.

\section{Related Work}
\label{sec:related}
The FaRO1~\cite{bolme2019faro,bolme2020face} framework arose from the need to create a convenient pipeline for biometric evaluations that would mitigate the often cumbersome integration tasks found in open-source and academic algorithms. The goal was an efficient framework in which components of a given pipeline (e.g., sensor source, template extractor, detector, matcher) could be swapped in and out with ease to provide the optimal system for a particular use case.  The initial development team was also concerned with scalability in the shift from cloud-based servers to edge-deployed systems.  To provide this optimized computational architecture, gRPC was tightly implemented to manage the streaming interfaces between server systems and client systems.

FaRO1 was utilized in the development of a specialized biometrics system for the identification of drivers and passengers in moving vehicles \cite{cornett2019through, ruby2020mertens}. Given FaRO1's successful performance in this implementation, it was further extended to incorporate more benchmarking and state-of-the-art algorithms to benefit the biometrics community and was released openly \cite{bolme2019faro, bolme2020face}.

In parallel to FaRO1's release as a gRPC server--client biometrics framework, other similar frameworks have also been developed. The Nvidia Triton Inference Server \cite{nvidia_triton_2022}, which began development in November 2018, provided similar gRPC client--server functionality but was developed with TensorRT specifically in mind. Additionally, FaRO2 now provides complete declarative infrastructure for chaining microservices together into workflows---a feature Triton does not have. 

Apart from Triton, other software suites aim to provide similar unified API function calls to hosted machine learning models.  Software such as Amazon's Sagemaker \cite{hudgeon_nichol_2020} provides cloud-specific hosting tools callable by a unified API, while libraries such as 
OpenVINO \cite{gorbachev2019openvino} make it possible to load models on many different hardware architectures. FaRO2 provides generic worker interfaces that harness APIs such as Sagemaker and OpenVINO, which allows for greater diversity of hostable models. In the next section, we will discuss the unique and novel aspects of FaRO2's architecture and implementation, and what sets it apart from other software in the model workflow space.

\section{FaRO2 Architecture Overview}
\label{sec:architecture}
FaRO1 was designed specifically as a unifying API harness for biometric algorithms, usually for the purposes of evaluations and experiments. The underpinning design decisions for FaRO2 revolve around the ability to quickly develop and deploy real-time, secure, and safe computer vision and biometric workflows in unconstrained settings, using a declarative paradigm that allows the user to easily connect a myriad of vision algorithms and microservices.

\begin{figure}
    \centering
    \includegraphics[scale=0.42]{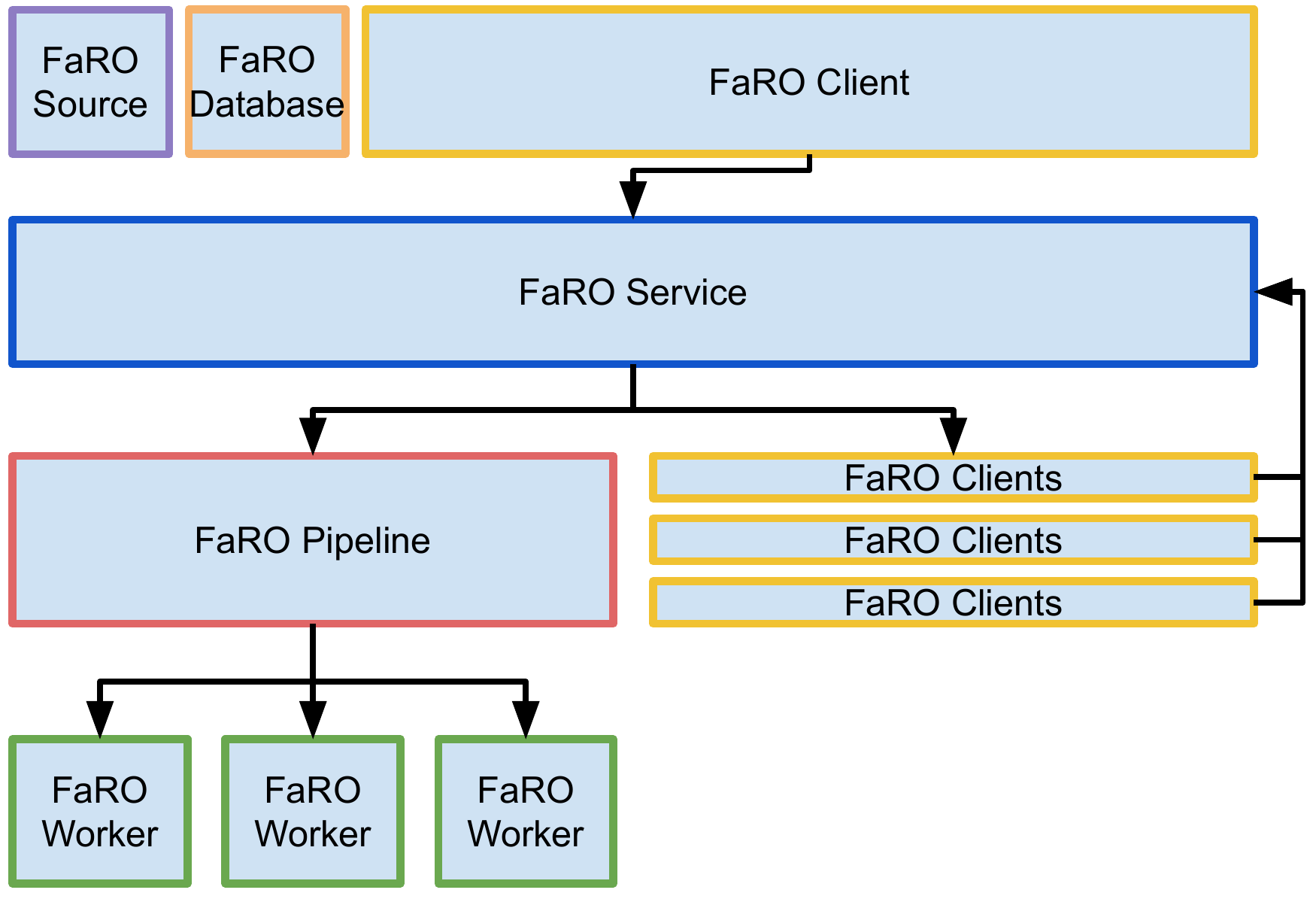}
    \caption{The hierarchy of FaRO entities. Generic gRPC messages called FaroRecord and FaroReply, which contain outputs from workers and pipelines, can be passed among all entities at any level. FaRO Clients connect to FaRO Services, which hold persistant instantiations of FaRO Pipelines. FaRO Pipelines contain chained-together graphs of FaRO Workers, which act as microservices. Each FaRO Service also contains persistent connections to other FaRO Clients in its area network, which in turn connect to their own services. This provides the basis for the recursive FaRO chaining that makes it so powerful.}
    \label{fig:faro_hierarchy}
\end{figure}

FaRO2 's design revolves around a gRPC server--client architecture that utilizes message passing and procedure calls to stream real-time video from client to server and asynchronously receive return results from server to client.  FaRO2 is built on three main hierarchical concepts: workers, pipelines, and services. The hierarchy itself can be seen in Figure \ref{fig:faro_hierarchy}. All of these entities perform machine learning tasks on inputs by accepting a generic gRPC message, called a FaroRecord, and returning a generic gRPC message, called a FaroReply. These record and reply messages act as a unifying language within FaRO and can be passed among between clients, servers, workers, and pipelines interchangeably. In this way, workers, pipelines, and services can pass messages either locally within themselves or between services and pipelines being hosted elsewhere.  As can be seen in Figure~\ref{fig:faro_hierarchy}, this implicitly creates a recursive hierarchical networking structure that allows workers hosted within remote services to propagate through network channel chains, similar to a mesh network. FaRO client calls can be made to connect these workers and pipelines in a variety of ways using a declarative interface.  These powerful concepts provide the foundation for FaRO2’s remarkable online configurability when being used to deploy and fine-tune distributed and heterogeneous machine learning systems in smart city infrastructures.

\subsection{FaRO Workers}
Each individual worker performs a single task, known as a microservice, such as detection, segmenting, or feature extraction. The worker contains the initialization constructors to load the particular libraries, models, and resources required to perform that task. Each worker also implements a set of initialization options. The FaroWorker also implements a method to report information about itself, including what type of microservice it provides, and what resources it has available to it. Each worker takes FaroRecords as input and returns FaroReplies as output. On an abstract level, all workers are able to interface with each other through records and replies but will throw exceptions if a given FaroReply does not contain the required input for a subsequent worker.

\subsection{FaRO Pipelines}
Each FaRO Pipeline consists of a directed acyclic graph workflow of FaRO Workers, chained together via their inputs and outputs. The FaRO Pipeline leverages the Multiprocessing library to allow for asynchronous calls to various workers in either an unordered manner or first-in-first-out order queues. In this way, local resources can be utilized to their fullest extent to perform parallel jobs that do not require order, or that require only minimal ordered dependencies between workers within the workflow.

The FaRO Pipeline itself subclasses the FaRO Worker and can utilize nested pipelines within its own workflow. This nested pipeline functionality helps provide further flexibility when creating more complex dependency graphs in distributed wide area networks. An example of this nested pipeline infrastructure is shown in Figure \ref{fig:faro_arch}.

\subsection{FaRO Services}
A FaRO Service implements the entire gRPC server required to make remote calls across a network channel. This server contains four main things: 
\begin{itemize}
    \item The code and infrastructure that allows online declaring of FaRO Pipelines constituted of chained-together FaRO Workers.
    \item A set of local FaRO Workers that are loadable and run-able within the FaRO Service's local environment.
    \item A set of FaRO Clients that all connect with other visible FaRO Services available on the network.
    \item One or more FaRO Databases that are endpoints to collect output from various FaRO Workers, in scenarios when databases must be created for enrollments or searches of various entities.
\end{itemize}

These main items allow users to connect to a FaRO Service via a FaRO Client, then create pipelines of FaRO Workers (either local or remote) via a declarative interface. These chains can be created from either fully local workers or a mixture of local and remote workers.

\subsection{FaRO Clients}
A FaRO Client acts as the interface that connects with any given FaRO Service. Each client connects to a single FaRO Service via a dedicated gRPC channel. Clients provide an API to call workers, pipelines within a FaRO Service, and a command--line interface (CLI) that allows end users to interact with the services and workers available to a given client.

\begin{figure}
    \centering
    \includegraphics[scale=0.43]{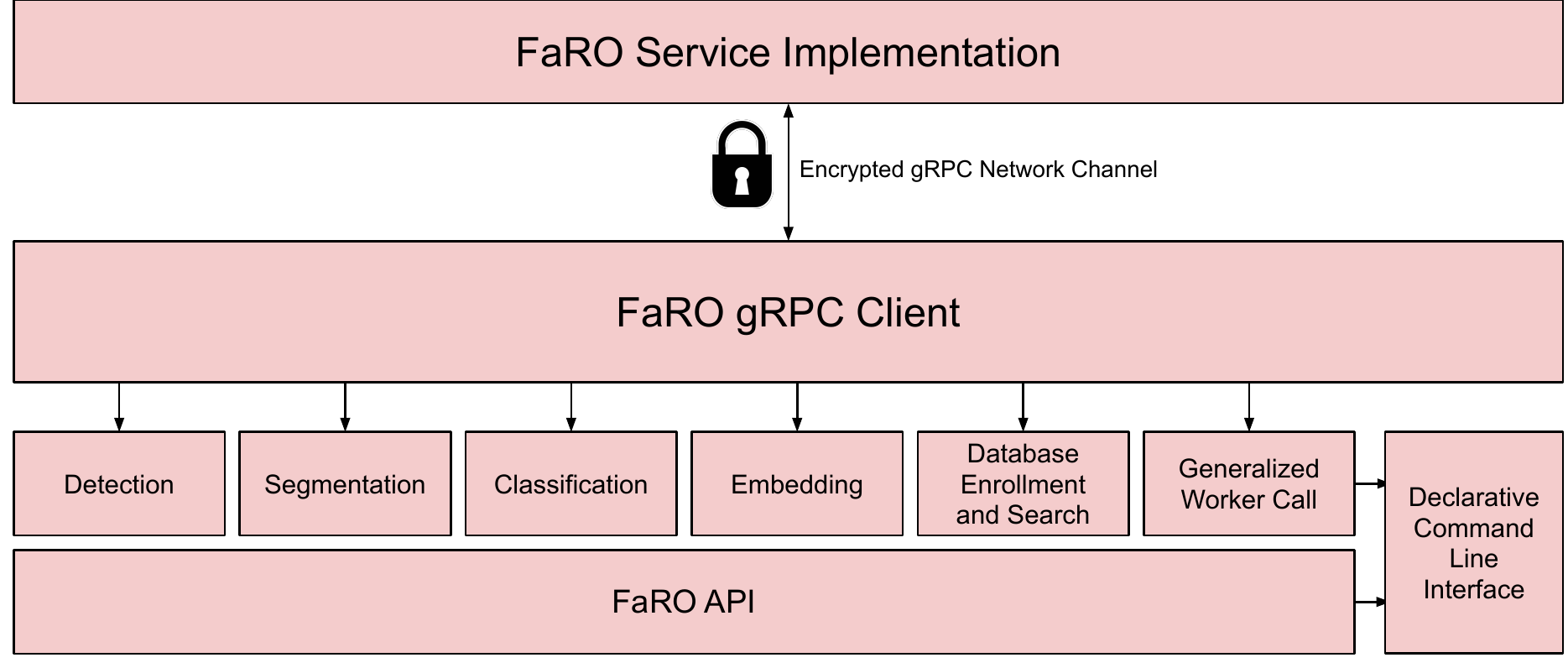}
    \caption{A visualization showing the specific and generic computer vision calls implemented in the FaRO Client that allow for communication to FaRO Services, and their nested workers and pipelines. The client provides a unified API and CLI to access workers and pipelines within a service. If a computer vision--type call that does not fall within the outlined client framework is required, such as a special type of superresolution or other processing, it can be implemented ad hoc using the ``generalized worker call.''}
    \label{fig:faro_client_struct}
\end{figure}

Clients have a set of specific, specialized remote procedure calls that can be requested from a FaRO Service, along with a generic call that can be made for procedures that do not fall within the general detect--extract--enroll--search architecture of computer vision. The specific remote procedure calls with dedicated implementations within the client are enumerated in Figure \ref{fig:faro_client_struct}.

Each client--service pair connects via an encrypted gRPC network channel. Section~\ref{sec:security} provides more detail on this channel encryption.

\subsection{FaRO Sources and Databases}
FaRO Sources implement a simple finite or indefinite iterator for streaming media.  Each FaRO Source implements a grab functionality, which ''grabs'' the next frame from a given camera or video file.  The built-in generic FaRO Source can read most video files and CODECs, along with the ability to read Real-Time Streaming Protocol, M3U8, and gstreamer syncs \cite{taymans2013gstreamer}. Other sources that utilize proprietary software development kits, such as Pylon or Vimba, can be easily implemented and integrated into the FaRO Ecosystem by subclassing the FaRO Source.

FaRO Clients connect directly to a FaRO Source and stream their output to connected services.  Because FaRO Clients and Services need not live within the same environment, software or environment conflicts can easily be resolved by hosting a service outside of the environment on which the camera or source must run. For example, if a given machine vision camera worked only with software designed for Windows, the FaRO Client could run within a Windows environment while streaming output to a service located on a DGX Linux environment suited for real-time processing.

FaRO Databases are designed to store enrolled templates and embeddings extracted from various worker microservices---at a high level, implementing record enrollment, deletion, and search. The built-in FaRO Database implementation also connects directly to the PHE layer to create a secure template storage solution that cannot leak private information. FaRO databases are loaded by as a persistent object within a FaRO Service and are, therefore, accessible as a remote database propagated through nested FaRO server--client connections.

\subsubsection{Zero-Configuration Networking}
FaRO utilizes Zero-Configuration Networking (ZeroConf)~\cite{stoica2002internet}, also known as Apple Bonjour, to make services discoverable within a local area network or wide area network. This feature creates a DNS-like service with which FaRO Services are easily addressable over the network, even in the presence of changing IP addresses. Services hosting both workers and pipelines broadcast their presence with a unique name. This allows other services within the network to automatically connect and discover other workers and pipelines available to it through service-to-client connections. Using ZeroConf, FaRO clients can discover and utilize all workers and pipelines active and visible on a given network without needing to know exact IP addresses.

\subsection{Streaming and Analytics in Real Time}
\begin{figure}
    \centering
    \includegraphics[scale=0.8]{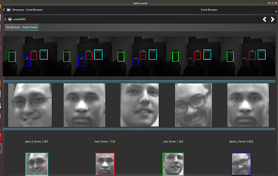}
    \caption{A simple cross-platform GUI built into FaRO for easy visualization of biometric data.}
    \label{fig:faro_gui}
\end{figure}

Because FaRO2 is designed as a real-time streaming service client framework, decisions were made with the end user in mind.  In Figure \ref{fig:faro_gui}, we show a simple graphical user interface (GUI), built directly into the FaRO2 library, that can display real-time streaming results from given workers and pipelines. This GUI can be deployed either on the server or client end, or both. While the GUI is relatively bare-bones, it is designed to be easily extensible and cross-platform capable. This feature provides users of FaRO2 either a out-of-the-box visualization tool for their real-time deployed systems or a straightforward guide on how to implement their own custom interface.

\subsection{Privacy and Network Security}
\label{sec:security}
\subsubsection{Partially Homomorphic Encryption Layer}

FaRO implements a built-in service that provides PHE using the Paillier Cryptosystem~\cite{o2008paillier}. More specifically, FaRO2 utilizes a modified version of the Python implementation of Paillier PHE provided here~\cite{PythonPaillier}. One of the bottlenecks of the Paillier Cryptosystem is the speed and efficiency with which the modular multiplicative inverse (MMI) can be run~\cite{hu2016analysis}. Traditionally, this calculation is performed utilizing the extended Euclidian algorithm (EEA)~\cite{naranjo2010applications}. However, naive implementations of the EEA can be prohibitively slow.

\begin{figure}
    \centering
    \includegraphics[scale=0.5]{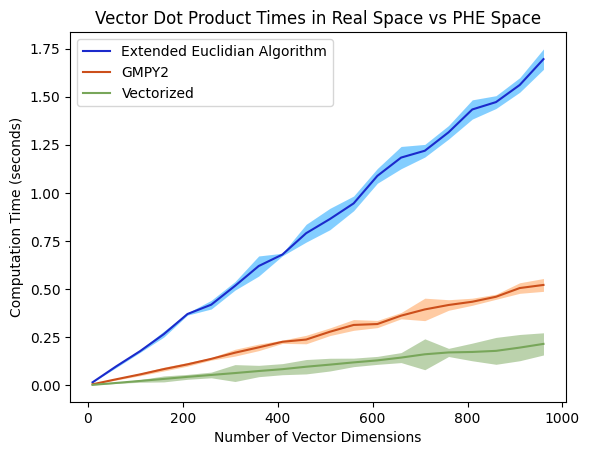}
    \caption{PHE encryption times for face vector embeddings of different dimensions. The original EEA, while the most straightforward, performs the slowest. We have markedly optimized the GNU implementation using GMPY2 (orange) to perform two to three times faster (blue) within the FaRO framework by utilizing vectorized operations to parallelize the EEA.}
    \label{fig:phe_times}
\end{figure}

In~\cite{PythonPaillier}, authors sped up this calculation for single numbers by utilizing the GMPY2 library~\cite{martelli}, a C-based Python interface that utilizes the GNU MPFR library for multiple-precision arithmetic~\cite{fousse2007mpfr}. For FaRO, we further vectorized this MMI using Numpy~\cite{projects_2021}. In Figure~\ref{fig:phe_times}, we show the effectiveness of this vectorized algorithm in the use of performing vector dot products for L1 distance calculations on face templates ranging from 1 to 1,024 dimensions. As can be seen in Figure~\ref{fig:phe_times}, the FaRO vectorized multiplicative inverse performs almost an order of magnitude faster than the original implementation in~\cite{martelli}.

\subsubsection{Encrypted gRPC Channels}
As a second layer of security, FaRO2 provides one-line flags to ensure that the gRPC channel connecting a FaRO Client and Service is encrypted using either RSA or ED25519. This allows data to be passed securely either on local gRPC channels or when being transmitted over wide area networks.

\section{Where to Find FaRO}
\label{sec:where}
FaRO2 is freely available and currently hosted on Github at \url{https://github.com/ORNL/faro/tree/FaRO2}. While currently hosted as a branch to FaRO1, FaRO2 will likely switch to its own repository sometime in the near future.

The repository contains documentation as well as Jupyter Notebooks to get going using examples of how to implement FaRO Workers, and how to utilize the FaRO2 API and CLI.

Currently, FaRO1 and FaRO2 are used on multiple projects "in the wild." A through-windshield imaging system~\cite{cornett2019through} utilizes the FaRO Framework to perform its recognition.  Similarly, the Deep-HDR fusion algorithm pipeline~\cite{ruby2020mertens} utilizes FaRO to route imagery from multiple camera sources to channels within the given network.

FaRO1 is also being utilized in driver safety projects~\cite{hooge2020evaluating} to determine what methods of data privacy work best to protect data of drivers in driver-facing camera systems. FaRO1 has also been effectively utilized for low-resource real-time computer vision on edge devices~\cite{bolme2022handheld,bolme2021rifle}.


\section{Acknowledgements and Copyright}
Notice:  This manuscript has been authored by UT-Battelle, LLC, under contract DE-AC05-00OR22725 with the US Department of Energy (DOE). The US government retains and the publisher, by accepting the article for publication, acknowledges that the US government retains a nonexclusive, paid-up, irrevocable, worldwide license to publish or reproduce the published form of this manuscript, or allow others to do so, for US government purposes. DOE will provide public access to these results of federally sponsored research in accordance with the DOE Public Access Plan (http://energy.gov/downloads/doe-public-access-plan).

Research sponsored by the Laboratory Directed Research and Development Program of Oak Ridge National Laboratory, managed by UT-Battelle, LLC, for the U. S. Department of Energy.

\section{Conclusions and Future Work}
\label{sec:conclusion}

FaRO2 is a ground-up reinvented declarative computer vision framework that builds upon the achievements and successes of FaRO1 by increasing flexibility and efficiency and responding to the growing need for truly secure, streaming-based inferencing frameworks for use in smart city and real-time surveillance implementations. 

A major avenue for future work is the continued development of the Oak Ridge Identity Testbed (ORID), a framework composed of sensors, biometric detection and identification algorithms, and computing platforms. Powered by FaRO2, ORID streams video data over the internal network from cameras deployed across Oak Ridge National Laboratory campus to one or more GPU servers for processing and routes results back out to create impactful visualizations for security operations centers and other demonstrations. It provides flexibility for rapid deployment of sensors and algorithms for testing and evaluation and the infrastructure required to build novel datasets with minimal lead time. Future work will also include integration of multimodal sensors and algorithms with FaRO2 and ORID, further investigation into privacy-protecting technologies, and automated collection and annotation of novel datasets for algorithm development.

{\small
\bibliographystyle{ieee_fullname}
\bibliography{egbib}
}

\end{document}